\documentclass[sigconf]{acmart}

\copyrightyear{2026}
\acmYear{2026}
\setcopyright{cc}
\setcctype{by}
\acmConference[CIKM '26]{Proceedings of the 35th ACM International Conference on Information and Knowledge Management}{November 07--11, 2026}{Rome, Italy}
\acmBooktitle{Proceedings of the 35th ACM International Conference on Information and Knowledge Management (CIKM '26), November 07--11, 2026, Rome, Italy}
\acmDOI{10.1145/3799682.3840206}
\acmISBN{979-8-4007-2539-5/2026/11}

\usepackage{graphicx}
\usepackage{booktabs}
\usepackage{tabularx}
\usepackage{array}
\title[Ishigaki-IDS-Bench]{Ishigaki-IDS-Bench: A Benchmark for Generating Information Delivery Specification from BIM Information Requirements}

\author{Ryo Kanazawa}
\affiliation{%
  \institution{ONESTRUCTION Inc.}
  \city{Tottori}
  \country{Japan}
}
\email{r.kanazawa@onestruction.com}
\author{Koyo Hidaka}
\affiliation{%
  \institution{ONESTRUCTION Inc.}
  \city{Tottori}
  \country{Japan}
}
\email{k.hidaka@onestruction.com}
\author{Teppei Miyamoto}
\affiliation{%
  \institution{ONESTRUCTION Inc.}
  \city{Tottori}
  \country{Japan}
}
\email{t.miyamoto@onestruction.com}
\author{Takayuki Kato}
\affiliation{%
  \institution{ONESTRUCTION Inc.}
  \city{Tottori}
  \country{Japan}
}
\email{t.kato@onestruction.com}
\author{Tomoki Ando}
\affiliation{%
  \institution{ONESTRUCTION Inc.}
  \city{Tottori}
  \country{Japan}
}
\email{t.ando@onestruction.com}
\author{Chenguang Wang}
\affiliation{%
  \institution{AWS GenAI Innovation Center}
  \city{Tokyo}
  \country{Japan}
}
\email{wangcg@amazon.co.jp}
\author{Dayuan Jiang}
\affiliation{%
  \institution{AWS GenAI Innovation Center}
  \city{Tokyo}
  \country{Japan}
}
\email{jiangdy@amazon.co.jp}
\author{Naofumi Fujita}
\affiliation{%
  \institution{ONESTRUCTION Inc.}
  \city{Tottori}
  \country{Japan}
}
\email{n.fujita@onestruction.com}
\author{Shuhei Saitoh}
\affiliation{%
  \institution{ONESTRUCTION Inc.}
  \city{Tottori}
  \country{Japan}
}
\email{s.saito@onestruction.com}
\author{Atomu Kondo}
\affiliation{%
  \institution{ONESTRUCTION Inc.}
  \city{Tottori}
  \country{Japan}
}
\email{a.kondo@onestruction.com}
\author{Koki Arakawa}
\affiliation{%
  \institution{ONESTRUCTION Inc.}
  \city{Tottori}
  \country{Japan}
}
\email{k.arakawa@onestruction.com}
\author{Daiho Nishioka}
\affiliation{%
  \institution{ONESTRUCTION Inc.}
  \city{Tottori}
  \country{Japan}
}
\email{d.nishioka@onestruction.com}

\renewcommand{\shortauthors}{Kanazawa et al.}

\begin{abstract}
Building Information Modeling (BIM) projects increasingly use Information Delivery Specification (IDS) to formalize information requirements in a machine-checkable XML format. Because IDS conditions are grounded in the Industry Foundation Classes (IFC) vocabulary, authoring them requires expertise in IFC concepts, validation tools, and property set conventions. Existing benchmarks for structured generation do not adequately capture the additional burden of vocabulary conformance and external-validator agreement that IDS imposes. We present Ishigaki-IDS-Bench, the first publicly released benchmark for IDS generation from BIM information requirements. The benchmark contains 166 examples spanning 83 practical scenarios authored in Japanese and English by six BIM/IDS experts, each paired with a gold IDS file and metadata covering input format, turn setting, target IFC versions, and construction domain. Evaluation proceeds in two stages: (i) formal validity scored by the buildingSMART IDSAuditTool along Processability, Structure, and Content, and (ii) content fidelity scored by facet-level macro-F1 against the gold IDS. Across 10 LLMs in zero-shot, the highest Facet F1 is 65.6\%, achieved by GPT-5.5, while the highest Content pass rate is only 33.1\%, achieved by Claude Opus 4.5. Ishigaki-IDS-Bench is released on Hugging Face (DOI 10.57967/hf/8873) under CC BY 4.0, and the evaluation code is released on Zenodo (DOI 10.5281/zenodo.20550510) under Apache-2.0.
\end{abstract}


\begin{document}


\maketitle

\section{Introduction}

Large language models (LLMs) are widely used to generate structured outputs such as JSON, SQL, and code\cite{willard2023outlines,beurerkellner2024guiding}. However, in practical domains, structured outputs are not sufficient simply because they are syntactically valid. They must simultaneously conform to industry-standard data formats, domain-specific vocabularies, version constraints, and audit results from external validation tools. Evaluation of structured generation that conforms to such domain standards remains less developed than evaluation of general-purpose JSON or SQL generation. This study addresses this problem using Information Delivery Specification (IDS), which describes information requirements in the architecture and construction domain in a machine-readable form.

Building Information Modeling (BIM) is an information foundation for handling building and infrastructure component types, materials, performance, and management information\cite{eastman2008bim}, and Industry Foundation Classes (IFC) is an international standard data format for sharing BIM models across different software systems\cite{iso16739}. Information Delivery Specification (IDS) is a standard specification that describes in XML ``which elements should have which information under which conditions'' for IFC models\cite{buildingsmartIDS}. For example, a requirement such as ``all walls shall have a fire-resistance rating, whose value is one of EI30, EI60, and EI90'' must map the IFC class for walls, the information item for fire-resistance rating, and the allowed-value constraint to IDS inspection units (facets).

In practice, information requirements are often written as natural-language specifications, tabular checklists, client requirements, meeting records, and similar documents. Creating IDS therefore requires expertise in BIM, IFC, and IDS, as well as the ability to interpret input documents. LLMs may support this transformation, but there is a lack of public benchmarks that jointly evaluate the formal validity of generated IDS, conformance to the IDS standard, consistency with IFC vocabulary, and content agreement with the input document.

We therefore propose Ishigaki-IDS-Bench, a benchmark for evaluating the ability to generate IDS~1.0-compliant XML from practical documents. The benchmark contains 166 examples obtained by expanding 83 practical scenarios into Japanese and English, corresponding gold IDS, and metadata for input format, language, turn setting, target IFC versions, and construction domain.

In addition, we design a two-stage protocol for evaluating generated IDS from both formal and content perspectives. In the first stage, IDSAuditTool\cite{buildingsmartIDSAudit} verifies IDS extractability, schema compliance, and conformance to the IDS standard and IFC vocabulary constraints. In the second stage, content agreement with the gold IDS is measured to capture generations that are formally valid but differ from the input document.

The contributions of this work are as follows.

\begin{itemize}
\item We construct an IDS generation benchmark based on practical use cases. Ishigaki-IDS-Bench contains 166 expert-created and verified examples, 83 scenarios, corresponding gold IDS, and multifaceted metadata.
\item We provide a two-stage evaluation protocol that combines formal-validity evaluation using IDSAuditTool with content-agreement evaluation against gold IDS.
\item We report zero-shot baselines for 10 LLMs and analyze successful cases and failure tendencies in IDS generation.
\end{itemize}

\section{Related Work}

Schema-constrained and grammar-constrained generation have been developed as methods for improving the formal validity of structured outputs generated by LLMs. Representative examples include methods that incorporate FSMs or CFGs into decoding\cite{willard2023outlines,geng2023grammar}, XGrammar, which co-designs the inference engine and grammar engine\cite{dong2025xgrammar}, Grammar-Aligned Decoding, which corrects distributional distortion caused by grammar constraints\cite{park2024grammarAligned}, and methods that use grammar masking for DSL generation\cite{netz2024grammarMasking}. These studies have mainly targeted outputs with general-purpose schemas or explicit syntactic constraints, such as JSON, SQL, code, and DSLs. In contrast, domain-specific XML standards such as IDS require not only syntactic validity but also consistency with IFC vocabulary, external standards, version constraints, property set conventions, and the judgments of validation tools. IDS generation should therefore be evaluated not merely as syntax-constrained generation, but as domain-standard-compliant structured generation.

For LLM evaluation in specialized domains, benchmarks have increasingly been constructed for areas that require expert knowledge, such as LawBench\cite{lawbench} and LeDQA\cite{liu2024ledqa} in the legal domain, EDINET-BENCH\cite{sugiura2025edinetbench} in the financial domain, and ECKGBench\cite{liu2025eckgbench} in the e-commerce domain. In the architecture and construction domain as well, studies have addressed building-code interpretation\cite{fuchs2024llmBuildingRegulations}, BIM compliance checking\cite{chen2024bimCompliance,madireddy2025llmCodeCompliance}, construction safety datasets\cite{ou2025buildingSaferSites}, BIM-GPT for applying LLMs to BIM information retrieval\cite{zheng2023bimgpt}, Qwen-BIM specialized for BIM design tasks\cite{lin2026qwenbim}, IFC-Agent for schema-guided multi-agent reasoning over IFC\cite{gao2026ifcagent}, MCP4IFC for editing IFC through code generation\cite{nithyanantham2025mcp4ifc}, IFC reasoning and editing\cite{tung2025ifcwhisperer}, and IDS/mvdXML standardization\cite{tomczak2022ids,tomczak2024ids,kladz2025idsbsdd,lee2020mvdmodules,son2022mvdxml}. However, most of these studies focus on question answering, retrieval, building-code interpretation, compliance judgment, BIM information retrieval, IFC model understanding, design support, and model editing. There is no public benchmark that directly generates IDS from BIM information requirements and integrates input documents, gold IDS, external-validator audits, and facet-level content-agreement evaluation. Ishigaki-IDS-Bench targets domain-standard XML generation that can be verified using an external validator and gold IDS, and is positioned at the intersection of schema-constrained generation and BIM/IFC-domain evaluation.

\section{Ishigaki-IDS-Bench}

\begin{figure}[t]
  \centering
  \includegraphics[width=\columnwidth]{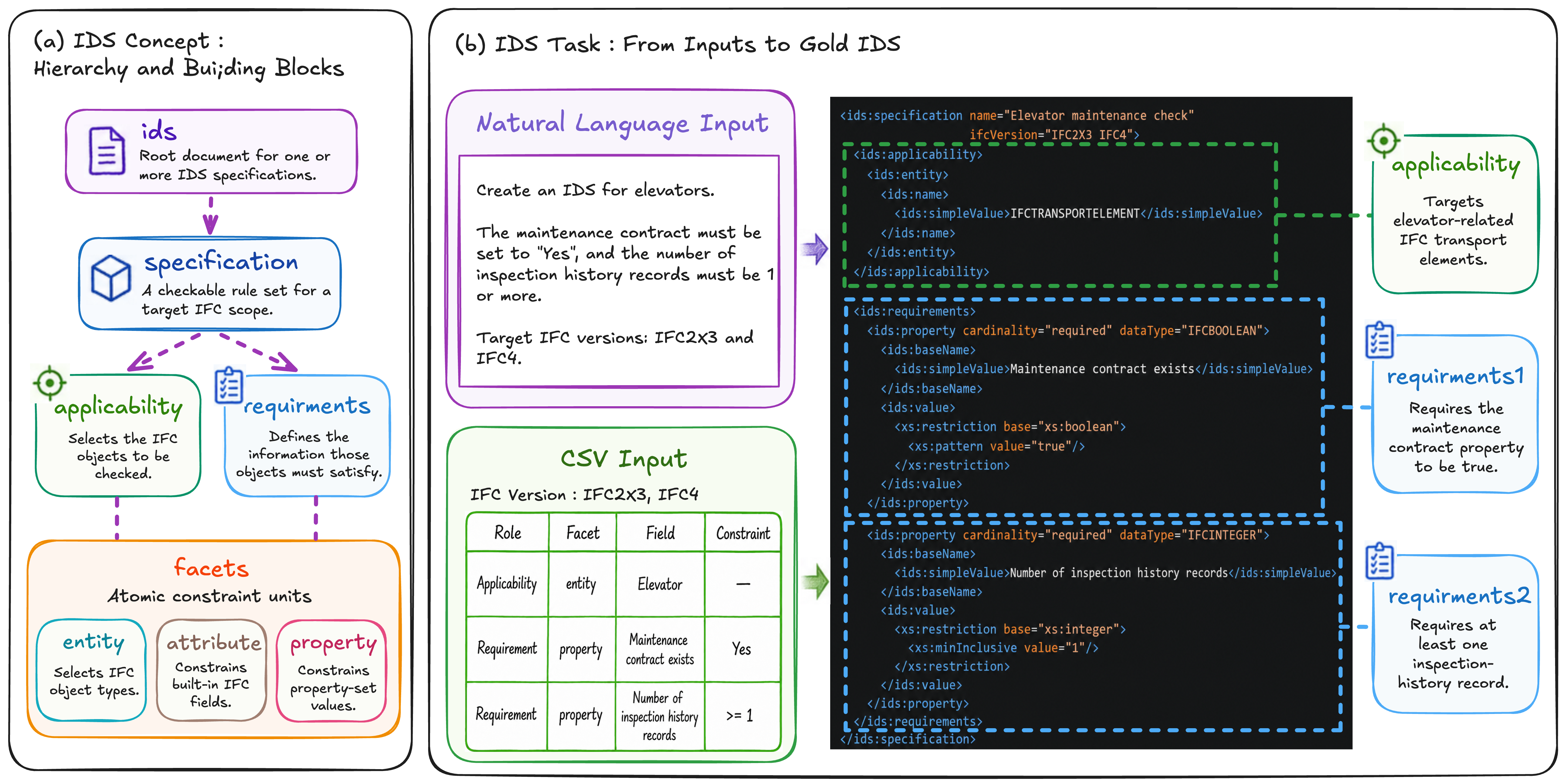}
  \caption{IDS hierarchy and task definition in Ishigaki-IDS-Bench. The left panel shows the main units that constitute IDS, and the right panel shows the correspondence used to create gold IDS from natural-language or CSV inputs.}
  \Description{Overview of the IDS hierarchy and the Ishigaki-IDS-Bench task from natural language or CSV inputs to gold IDS.}
  \label{fig:ids-task}
\end{figure}

\begin{figure*}[!t]
  \centering
  \includegraphics[width=0.8\textwidth]{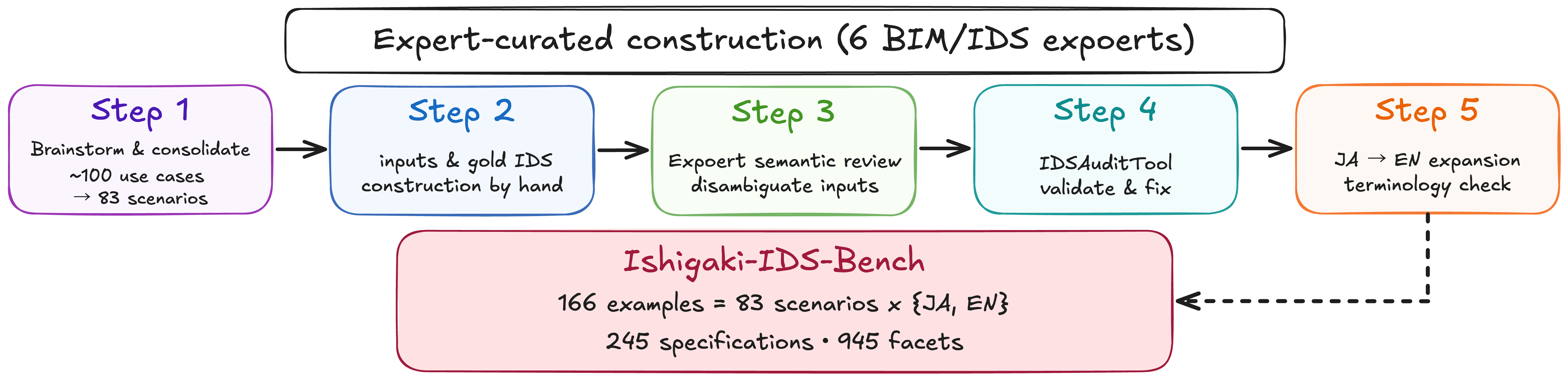}
  \caption{Construction pipeline of Ishigaki-IDS-Bench. Six BIM/IDS experts consolidated approximately 100 candidates into 83 scenarios, conducted gold IDS authoring, semantic review, IDSAuditTool validation, and JA/EN expansion. The artifacts include 166 examples, 245 specifications, and 945 facets.}
  \Description{Construction pipeline of Ishigaki-IDS-Bench from expert brainstorming and gold IDS authoring to audit validation and bilingual expansion.}
  \label{fig:dataset-construction}
\end{figure*}

\paragraph{Task and scope.}
Ishigaki-IDS-Bench targets the task of generating complete IDS files that conform to IDS~1.0 from information requirements. Each input includes a requirement written in CSV or natural language, an output file name, and one or more target IFC versions. The model outputs the corresponding IDS using only the requirements explicitly stated in the input.

Figure~\ref{fig:ids-task} shows the IDS structure and representative input--output pairs. IDS consists of one or more \texttt{specification} elements, each with \texttt{applicability} for the target and \texttt{requirements} for required conditions. These conditions are represented as facets. We evaluate \texttt{entity}, \texttt{attribute}, and \texttt{property} facets, which represent IFC element types, basic IFC-schema fields, and practical additional information. Thus, the task is not XML formatting but a structured generation task that maps practical information requirements to IDS target and requirement conditions. Because this study focuses on target specification and information-item specification as the core of practical information requirements, \texttt{classification}, \texttt{material}, and \texttt{partOf} are excluded from evaluation.

\paragraph{Taxonomy and statistics.}
Table~\ref{tab:dataset-stats} summarizes the five-axis taxonomy over 166 examples. The benchmark expands 83 scenarios into Japanese and English examples. An example is one input context paired with one gold IDS. For the 44 multi-turn examples, the model receives a fixed history containing the initial user request, dataset-provided assistant IDS drafts, and subsequent user revision requests. The model is required to produce a complete IDS reflecting the final revision request, while the final assistant IDS is withheld as the gold reference. A scenario is the base use case that groups corresponding examples representing the same information requirement. Each example is annotated with input format, language, turn setting, target IFC versions, and construction domain, enabling analysis of performance differences between natural-language and CSV inputs, Japanese and English inputs, single-turn and multi-turn settings, target IFC versions, and construction domains.

\begin{table}[t]
  \centering
  \small
  \setlength{\tabcolsep}{6pt}
  \renewcommand{\arraystretch}{1.08}
  \caption{Metadata distribution over 166 Ishigaki-IDS-Bench examples; IFC-version counts exceed 166 because an example may target multiple versions.}
  \label{tab:dataset-stats}
  \begin{tabular}{@{}ll@{}}
    \toprule
    Axis & Distribution \\
    \midrule
    Input format & CSV=58, Natural language=108 \\
    Language & JA=83, EN=83 \\
    Turn setting & Single-turn=122, Multi-turn=44 \\
    Target IFC versions & IFC2X3=78, IFC4=54, IFC4X3=70 \\
    Domain & Architecture=48, Structural=42, MEP=40, General=36 \\
    \bottomrule
  \end{tabular}
\end{table}

\begin{table*}[t]
  \centering
  \footnotesize
  \setlength{\tabcolsep}{4pt}
  \renewcommand{\arraystretch}{1.08}
  \caption{Zero-shot baseline results on Ishigaki-IDS-Bench. Model groups are divided based on whether their weights are publicly available. Processability, Structure, and Content denote first-stage audit rates, and Facet F1, Recall, and Precision denote second-stage macro averages over 166 examples.}
  \label{tab:baseline-results}
  \begin{tabular}{@{}lrrrrrr@{}}
    \toprule
    & \multicolumn{3}{c}{Stage 1} & \multicolumn{3}{c}{Stage 2 (Facet scorer)} \\
    \cmidrule(lr){2-4}\cmidrule(l){5-7}
    Model & Processability & Structure & Content & Facet F1 & Recall & Precision \\
    \midrule
    \multicolumn{7}{@{}l}{\textit{Closed proprietary LLMs}} \\
    GPT-5.5 (xhigh) & 94.6 & 30.1 & 27.7 & \textbf{65.6} & \textbf{65.8} & \textbf{65.6} \\
    Claude Opus 4.5 & \textbf{99.4} & \textbf{45.8} & \textbf{33.1} & 42.2 & 45.5 & 40.5 \\
    Gemini 3.1 Pro & 97.0 & 37.3 & 25.9 & 52.2 & 52.6 & 52.2 \\
    \midrule
    \multicolumn{7}{@{}l}{\textit{Open-weight LLMs}} \\
    gpt-oss-120b (xhigh) & 98.8 & 13.9 & 12.0 & 21.6 & 21.4 & 21.9 \\
    Kimi K2.6 & \textbf{99.4} & 41.0 & 22.3 & 48.4 & 48.9 & 48.4 \\
    DeepSeek V4 Pro & 97.0 & 19.3 & 16.3 & 40.0 & 36.3 & 49.1 \\
    Qwen3.5-397B-A17B & 98.2 & 20.5 & 19.3 & 26.8 & 26.2 & 28.5 \\
    Qwen3-32B & 37.3 & 13.9 & 10.2 & 22.3 & 21.7 & 23.5 \\
    Qwen3-14B & 96.4 & 18.1 & 13.9 & 21.3 & 21.2 & 21.6 \\
    Qwen3-8B & 26.5 & 20.5 & 15.7 & 20.2 & 20.0 & 20.5 \\
    \bottomrule
  \end{tabular}
  \par\vspace{2pt}
  {\footnotesize\raggedright\textit{Note.} Decoding settings were fixed for each model family before evaluation and were not tuned on Ishigaki-IDS-Bench. Qwen-family models were evaluated with \texttt{temperature=0.6}, \texttt{top\_p=0.95}, \texttt{top\_k=20}, \texttt{min\_p=0.0}, and other models with \texttt{temperature=0.0}, \texttt{top\_p=1.0}; the output length limit for all models was 15,000 tokens. The \texttt{(xhigh)} notation indicates a setting in which reasoning effort was fixed to \texttt{xhigh}. Truncation, structure-constrained decoding, repair, and regeneration were not used.\par}
\end{table*}

\paragraph{Dataset construction.}
Because public IDS examples are limited, we prioritized expert-judged practical representativeness over mechanical IFC-schema coverage. Six BIM/IDS experts, five with buildingSMART-related experience and five with IFC/IDS practice, consolidated about 100 candidates into 83 scenarios and manually authored input information requirements and gold IDS through the procedure shown in Figure~\ref{fig:dataset-construction}. Quality was checked through semantic review, IDSAuditTool validation, facet-scorer reruns, and JA/EN terminology checks; two additional ML experts reviewed the task definition, metadata, evaluation procedure, and reproducibility. Because the gold IDS files were created through expert authoring and standards-compliance validation rather than independent parallel annotation by multiple annotators, pairwise IAA was not computed. The scale of 166 examples results from prioritizing expert-validated, standards-compliant samples over synthetic augmentation.

\section{Evaluation Protocol}

Evaluation on Ishigaki-IDS-Bench consists of two stages: formal validity evaluation using IDSAuditTool (ids-tool v1.0.96) and facet-level content-matching evaluation against gold IDS. In the first stage, Processability reports whether an IDS candidate can be supplied to the audit tool, Structure reports XML and IDS XSD validity, and Content reports compliance with the IDS standard and IFC vocabulary constraints. In the second stage, we use a facet scorer that compares gold IDS and generated IDS. The comparison targets are IFC-version labels, entity, attribute, property, and the dataType/cardinality of property; exact matching is computed over a multiset of comparison blocks consisting of applicability/requirements, facet type, and constraint content. For IDS files containing multiple specifications, specifications in the gold IDS and generated IDS are aligned, and only the comparison representation is canonicalized for XML namespaces, constraint child elements, and simpleValue ordering. Precision, recall, and F1 are macro-averaged over examples, and LLM-as-judge is not used because direct comparison against gold IDS is possible. Because practical requirements may omit property set names, we consistently apply the non-IDS-standard internal convention explicitly stated in the zero-shot prompt to all models, and publish the corresponding slice and convention together with the evaluation scripts.

\section{Baseline Evaluation}

\paragraph{Experimental setup.}
We evaluated the 10 LLMs in Table~\ref{tab:baseline-results} in a zero-shot setting in May 2026. All models receive the same system prompt and the example-specific context containing the information requirements and target IFC versions. No separate single-turn or multi-turn examples, IDS outputs, or gold IDS are added as demonstrations. Prior IDS drafts in a multi-turn example are part of that example's fixed context, not demonstrations. If no IDS candidate could be extracted from the output or XML parsing failed, the output was treated as invalid. We did not perform multiple-sample selection, reranking, post-generation repair, regeneration, or manual correction.

\paragraph{Results and findings.}
Table~\ref{tab:baseline-results} shows the results. Even the model with the highest Facet F1 score, GPT-5.5, reaches only 65.6\% Facet F1 and a 27.7\% Content pass rate, indicating that XML generation that complies with the IDS standard and IFC vocabulary constraints remains difficult even when some input requirements are recovered as facets. High Processability with low Content is common, exposing a gap between audit-tool processability and standards-compliant content generation.

In Stage 2, GPT-5.5 and Gemini 3.1 Pro lead facet-level agreement, while Kimi K2.6 lags in Content and Facet F1 despite high Processability. Within the Qwen family, larger models show broadly higher Facet F1 but non-monotonic audit pass rates, indicating that scale alone does not ensure robust IDS processability or standards-compliant content generation.

For GPT-5.5, the 44 multi-turn examples reached Content=79.5\% and Facet F1=84.8\%, compared with 9.0\% and 58.6\% on the 122 single-turn examples. This pattern suggests that history-conditioned IDS revision may be more tractable than generation from scratch in this benchmark. CSV inputs achieved higher Facet F1 than natural-language inputs (77.3\% vs. 59.2\%).

Qualitatively, we observed three failure tendencies. First, many outputs can be extracted and audited as IDS candidates but still violate the IDS XSD, IDS standard, or IFC vocabulary constraints, appearing as lower Structure or Content scores. Second, \texttt{entity} and \texttt{attribute} facets are relatively easy to recover, whereas errors are common in \texttt{property} facets involving property set names, cardinality, and value constraints. Third, some natural-language errors involved ambiguous mappings between requirement targets and value constraints, leading to facet omissions or overgeneration.

\section{Availability, Ethics, and Maintenance}

The dataset and gold IDS, with per-example IDs and five-axis metadata, are released on Hugging Face under CC BY 4.0 (dataset DOI 10.57967/hf/8873)~\cite{ishigaki_ids_bench_dataset}; the evaluation scripts and zero-shot prompt are released on GitHub/Zenodo under Apache-2.0 (fixed-release DOI 10.5281/zenodo.20550510)~\cite{ishigaki_ids_bench_software}. The dataset excludes personal information, real IFC files, confidential project data, and copyrighted real-project documents; releases will be versioned and maintained on Hugging Face, Zenodo, and GitHub.

\section{Limitations}

Ishigaki-IDS-Bench has scope limitations. This study focuses on the \texttt{entity}, \texttt{attribute}, and \texttt{property} facets, and leaves \texttt{classification}, \texttt{material}, and \texttt{partOf} for future extension. Although the inputs are based on practical use cases, end-to-end performance on noisy real document collections is not directly measured because confidential documents and real IFC models are not redistributed. The handling of unspecified property sets is an internal evaluation convention rather than part of the IDS standard. The 166 examples are intended for diagnosis rather than large-scale training, and few-shot prompting, retrieval-augmented prompting, fine-tuning, and grammar-constrained decoding remain subjects for future comparison.

\section{Conclusion}

Ishigaki-IDS-Bench is an expert-created and verified public benchmark for IDS~1.0-compliant XML generation from BIM information requirements. Across 10 zero-shot baselines, the highest Facet F1 (65.6\%) and the highest Content pass rate (33.1\%) were achieved by different models, showing that IDS generation requires separate evaluation of domain vocabulary, standards compliance, and facet-level content consistency, not only syntactic well-formedness.

\section*{Acknowledgments}

This work was conducted as part of the GENIAC (Generative AI Accelerator Challenge) Project, which aims to strengthen Japan’s capability to develop generative AI and is promoted by the Ministry of Economy, Trade and Industry (METI) and the New Energy and Industrial Technology Development Organization (NEDO).

\section*{GenAI Usage Disclosure}

Generative AI tools were used as part of the research subject and experimental procedure. Specifically, zero-shot generation results were obtained from each LLM and used as baseline evaluation targets on Ishigaki-IDS-Bench. The input examples, metadata, gold IDS, evaluation design, metric computation, and result interpretation for Ishigaki-IDS-Bench were created and checked by the authors and experts. LLMs were not used to construct the benchmark examples or gold IDS files. During manuscript writing, generative AI tools including ChatGPT and DeepL were used for translation, language polishing, terminology consistency checks, and LaTeX formatting assistance. The authors checked and revised the outputs and take responsibility for all claims, experimental design, analysis, conclusions, and the final manuscript.

\bibliographystyle{ACM-Reference-Format}
\bibliography{references}

\end{document}